\documentclass[pdflatex, sn-mathphys-num]{sn-jnl}

\usepackage{graphicx}%
\usepackage{multirow}%
\usepackage{amsmath,amssymb,amsfonts}%
\usepackage{amsthm}%
\usepackage{mathrsfs}%
\usepackage[title]{appendix}%
\usepackage{xcolor}%
\usepackage{textcomp}%
\usepackage{manyfoot}%
\usepackage{booktabs}%
\usepackage{algorithm}%
\usepackage{algorithmicx}%
\usepackage{algpseudocode}%
\usepackage{listings}%
\usepackage{color}

\theoremstyle{thmstyleone}%
\theoremstyle{thmstyletwo}%

\theoremstyle{thmstylethree}%

\begin{document}

\title[Article Title]{Physical Data Embedding for Memory Efficient AI}


\author[1]{\fnm{Callen} \sur{MacPhee}}

\author[1]{\fnm{Yiming} \sur{Zhou}}

\author[1]{\fnm{Bahram} \sur{Jalali}}

\affil[1]{\orgdiv{Electrical and Computer Engineering Department}, \orgname{University of California, Los Angeles}, \orgaddress{\street{420 Westwood Plaza}, \city{Los Angeles}, \postcode{90095}, \state{CA}, \country{United States}}}

\nocite{zhou2024nonlinear}

\abstract{
Deep neural networks (DNNs) have achieved exceptional performance across various fields by learning complex, nonlinear mappings from large-scale datasets. However, they face challenges such as high memory requirements and computational costs with limited interpretability. This paper introduces an approach where master equations of physics are converted into multilayered networks that are trained via backpropagation. The resulting general-purpose model effectively encodes data in the properties of the underlying physical system. In contrast to existing methods wherein a trained neural network is used as a computationally efficient alternative for solving physical equations, our approach directly treats physics equations as trainable models. We demonstrate this physical embedding concept with the Nonlinear Schrödinger Equation (NLSE), which acts as \textit{trainable} architecture for learning complex patterns including nonlinear mappings and memory effects from data. The network embeds data representation in orders of magnitude fewer parameters than conventional neural networks when tested on time series data. Notably, the trained ``Nonlinear Schrödinger Network'' is interpretable, with all parameters having physical meanings. This interpretability offers insight into the underlying dynamics of the system that produced the data. The proposed method of replacing traditional DNN feature learning architectures with physical equations is also extended to the Gross-Pitaevskii Equation, demonstrating the broad applicability of the framework to other master equations of physics. Among our results, an ablation study quantifies the relative importance of physical terms such as dispersion, nonlinearity, and potential energy for classification accuracy. We also outline the limitations of this approach as it relates to generalizability.} 



\hbadness=10000 

\maketitle

\section{Introduction}\label{sec1}
Deep neural networks (DNNs) have emerged as a powerful tool for learning complex, nonlinear mappings from data, achieving state-of-the-art performance in various fields including computer vision, natural language processing, and scientific computing. By leveraging their deep hierarchical structure along with billions or even trillions of trainable parameters, DNNs have demonstrated remarkable capability to learn complex patterns from vast amounts of data.  However, the success of DNNs comes with challenges, such as reliance on large-scale datasets, high computational and memory costs, and limited interpretability. 

As DNN models grow in size and complexity, the memory wall, a fundamental bottleneck in computing architectures, has become increasingly problematic ~\cite{rajbhandari2021zero}. In the case of AI data centers, as model parameter counts have increased from millions to hundreds of billions, model memory requirements have severely outpaced on-chip capacity ~\cite{gholami2024ai}. This has lead to the adoption of power-hungry solutions such as memory disaggregation and optically connected memory ~\cite{gonzalez2022optically}. Memory bandwidth rather than computational throughput has become a limiting factor, with direct implications for power consumption ~\cite{yoo2021prospects}.

The economic ramifications of this memory wall extend beyond mere technical inefficiency. Data center operators face escalating cooling requirements, increased infrastructure costs, and diminishing returns on computational investments ~\cite{cheng2021survey}. The capital expenditure necessary to scale infrastructure for larger models grows non-linearly with model size, challenging the sustainability of continued parameter scaling as a path to improved model performance ~\cite{wu2022sustainable}.

Additionally, the surge in model size further obscures the inner workings of DNNs, undermining interpretability. It is becoming increasingly difficult to trace decisions back to salient features or transparent intermediate representations ~\cite{molnar2020interpretable, lipton2018mythos}, posing significant challenges in domains where explainability is crucial, such as healthcare and finance. Despite ongoing efforts in explainable artificial intelligence, the growing complexity of large-scale models continues to hamper interpretability techniques, emphasizing the need for research into new architectures or methodologies that strike a balance between model capacity and transparency ~\cite{rudin2019stop}.

\subsection{Related Work in Physics-AI Symbiosis}

To address these challenges, there is a growing interest in hybrid approaches that integrate physics with AI to enable greater efficiency in language, vision, control and more~\cite{karniadakis2021physics, jalali2022physics, carleo2019machine, rai2020driven}. In contrast to predominantly high-dimensional empirical methodologies in AI, the remarkable success of physics in explaining nature stems from its reliance on low-dimensional, deterministic models that precisely characterize a wide array of phenomena. The design principles that underlie these physical equations can offer strategies for creating more parameter-efficient, interpretable AI approaches. For example, physical principles can be leveraged as prior knowledge to design neural networks with special architectures and loss functions, enabling more efficient modeling ~\cite{raissi2019physics, li2020fourier, hamerly2019large}. Moreover, physical systems can act as analog computers to perform specialized computations or transform data in a manner that reduces the burden on digital processors~\cite{lin2018all, mohammadi2019inverse, zhou2022nonlinear, haensch2018next}

An emerging trend in blending physics and AI is the use of neural networks as computationally efficient surrogates for solving partial differential equations (PDEs). This approach has the potential to significantly accelerate forward simulations and inverse design used in the optimization of engineered physical systems. For example, neural networks can compute the transmission spectra of metasurfaces orders of magnitude faster than numerically solving the full Maxwell's equations~\cite{liu2018generative, lim2022maxwellnet}. Similarly, neural networks trained on real-world data can model intricate fluid flows for applications like weather forecasting and aerodynamics, circumventing the need to directly solve the Navier-Stokes equations~\cite{kutz2017deep}. More ambitious attempts have been made to develop general neural network architectures that can learn different types of PDEs. The physics-informed neural network (PINN)~\cite{raissi2019physics} is an example which utilizes a physics-driven loss derived from the equation itself to solve one instance of the PDE with a fixed set of parameters and conditions. Fourier Neural Operator (FNO)~\cite{li2020fourier} takes a step further by learning the nonlinear mapping of an entire family of PDEs.

\section{Physical Embedding with the Nonlinear Schrödinger Equation}

Although these approaches hybridize physics and AI, they do not necessary address the core challenges of parameter-efficiency and interpretibility. In contrast, this work proposes the concept of \textit{physical embedding}, wherein data representations are encoded in the physical properties of a medium rather than distributed across neural network parameters. Unlike existing works which use neural networks to replace PDEs by approximating their solutions ~\cite{raissi2019physics,li2020fourier}, this approach trains a PDE to replace a neural network by directly learning the PDE's material-dependent coefficients. By leveraging the low-dimensional nature of physical laws, such physical embedding represents complex data with significantly fewer parameters while achieving accuracy comparable to traditional neural networks on nonlinear classification tasks. 

In demonstration, we introduce the ``Nonlinear Schrödinger Network'' (NSN), a novel physics-inspired architecture that operates entirely in the digital domain. This model treats the Nonlinear Schrödinger Equation (NLSE) as a \textit{trainable} model to learn complex patterns from data. By optimizing the coefficients of the NLSE using backpropagation and stochastic gradient descent (SGD), nonlinear mappings are embedded in the physical parameters of the equation, which transforms the data adaptively for AI tasks. 

\begin{figure*}
    \centering
    \includegraphics[width=5in]{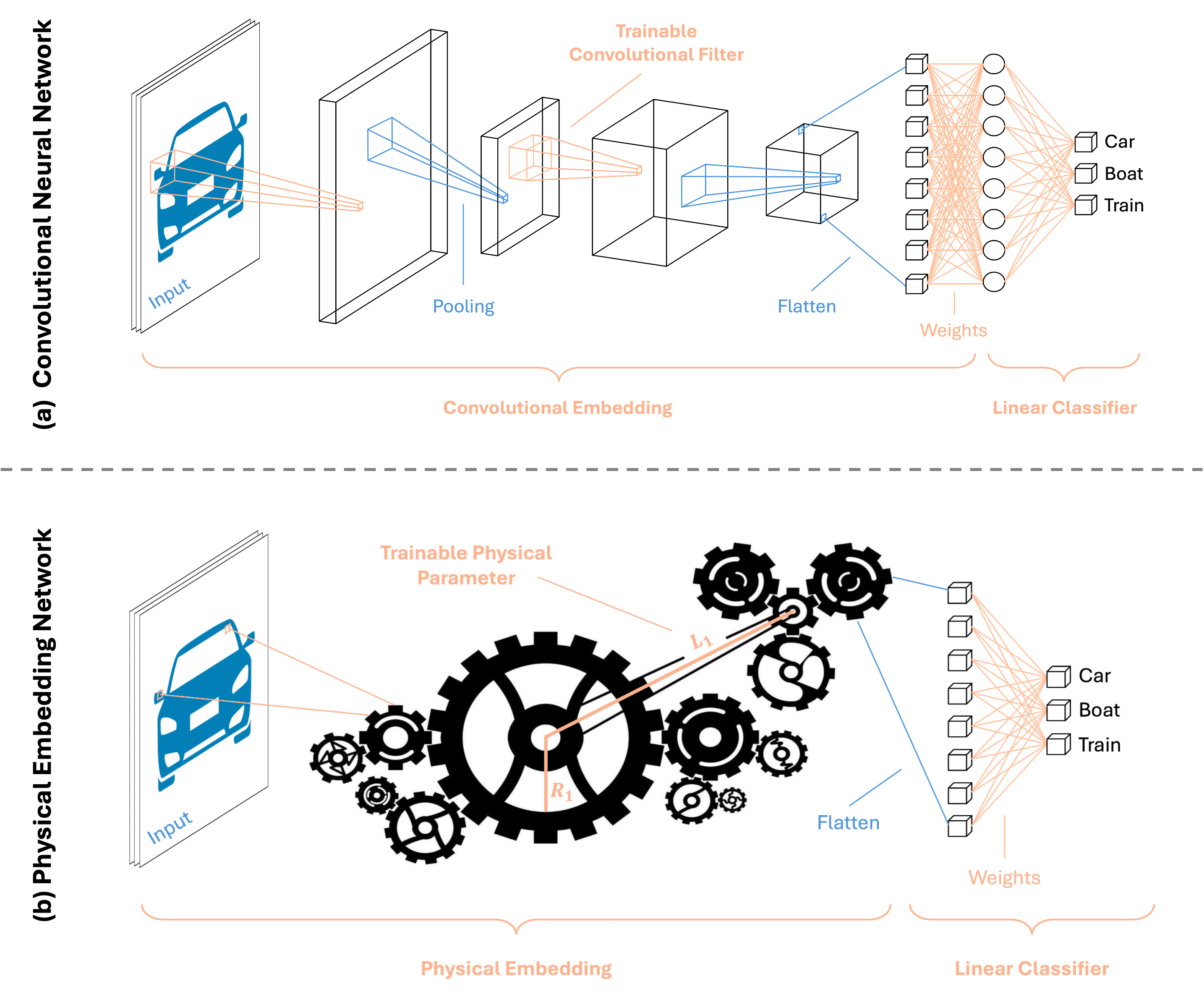}
    \caption{Comparison of Traditional Convolutional Feature Learning with Physical Embedding. Trainable steps are shown in orange. (a) \textbf{Convolutional Neural Network} with traditional convolutional feature learning layers followed by flattening, nonlinear activation, and linear classifier. (b) \textbf{Physical Embedding Network} wherein a PDE replaces the convolutional layers of a neural network. This embedding utilizes the PDE's characteristics as the parameters for a deep learning framework. The physical properties of the system are used as learnable parameters, which preserves the elegance and efficiency inherent to the underlying physics. This effectively embeds the complex patterns and relationships of the data into a handful of physically meaningful properties, such as the radius of and distance between gears, in this analogy. The transformed data can then be efficiently separated by a linear classifier.}
    \label{fig:physical_embedding}
\end{figure*}

\subsection{Physical Embedding}

In traditional neural networks, information is encoded within millions or billions of parameters distributed across a general network architecture ~\cite{gurney2018introduction}. By composing many simple, trainable operations, these models capture patterns, relationships, and abstractions from the data, essentially embedding representation in the parameters of the network \cite{bengio2013representation}.

The concept of physical embedding, introduced here, represents a fundamentally different approach. Rather than learning in an extensive parameter space, the approach trains the \textit{physical properties} of system governed by a PDE. This effectively embeds the complex patterns and relationships of the data into a handful of physically meaningful properties that describe the system material-dependent dynamics, with the transformation of the data handled by the underlying mechanics of the PDE. As such, the high-dimensional embedding layers of DNNs can be completely replaced by solving the trained PDE for a given input. This process is remarkably parameter efficient, requiring orders of magnitude fewer parameters than conventional neural networks while maintaining comparable performance. Additionally, since the embedding space is constrained by physical laws, the overall learned transformation is interpretable via the original physics through which the embedding equation was designed. Each parameter has a clear physical meaning, offering insight into how the data is being transformed at each layer of the network. One further advantage of this approach is its potential for analog implementation, which can drastically reduce latency compared to traditional digital systems. 

In essence, this reframes the problem of representation learning to one of optimizing the coefficients of a predetermined, interpretable physical system. This process bridges the gap between data-driven machine learning and physics-based modeling, offering a more interpretable and parameter-efficient alternative to traditional neural networks. To implement this paradigm in practice, we turn to the Nonlinear Schrödinger Equation (NLSE), a canonical equation in physics that naturally captures a rich variety of nonlinear behaviors and memory effects.

\subsection{The Nonlinear Schrödinger Equation as a Neural Network}

The Nonlinear Schrödinger Equation (NLSE) is one of the master equations of optics that appears in many areas of mathematical physics and applied science. A nonlinear partial differential equation, it accounts for three primary physical effects: attenuation, dispersion, and nonlinearity. These effects, in the limit of constant group delay dispersion (GDD), are represented by parameters $\alpha$, $\beta_2$, and $\gamma$, respectively, as illustrated in Equation \ref{eq:nlse}. 

\begin{equation}
    \frac{\partial E(t,z)}{\partial z} = -\frac{\alpha}{2} E(t,z) - i\frac{\beta_2}{2} \frac{\partial^2 E(t,z)}{\partial t^2} + i\gamma |E(t,z)|^2 E(t,z)
    \label{eq:nlse}
\end{equation}
\newline

Notably, these effects mirror core computational functions in deep neural networks, representing efficient alternatives grounded in fundamental physical processes. Physically, dispersion describes how different spectral components of a wave travel at different velocities, producing pulse reshaping over propagation. From a dynamical systems viewpoint, this introduces a memory effect: the present wave profile depends not only its current amplitude but also on its past configurations. In a neural network analogy, dispersion serves as a proxy for memory–or a form of recurrent context or state–allowing information to mix and influence over space and time. Attenuation captures the loss or dissipation of energy over distance, causing the overall amplitude of the wave to decrease. Numerically, in a neural-network-like framework, this effect parallels a gating mechanism: certain features are scaled down as they travel across layers (propagation steps). Nonlinearity governs how the wave interacts with itself, giving rise to effects such as self-phase modulation or self-focusing. This self-interaction means that the wave’s amplitude can alter its own speed and phase, thereby coupling the wave’s magnitude and phase in a nontrivial way. Computationally, nonlinearity provides the key to representing and learning nonlinear relationships. Like nonlinear activation functions in neural networks, it enables the model to capture richer, more intricate behavior than could be obtained with a purely linear system. Together, these physically grounded mechanisms can be harnessed to learn complex transformations.

\begin{figure*}
    \centering
    \includegraphics[width=5in]{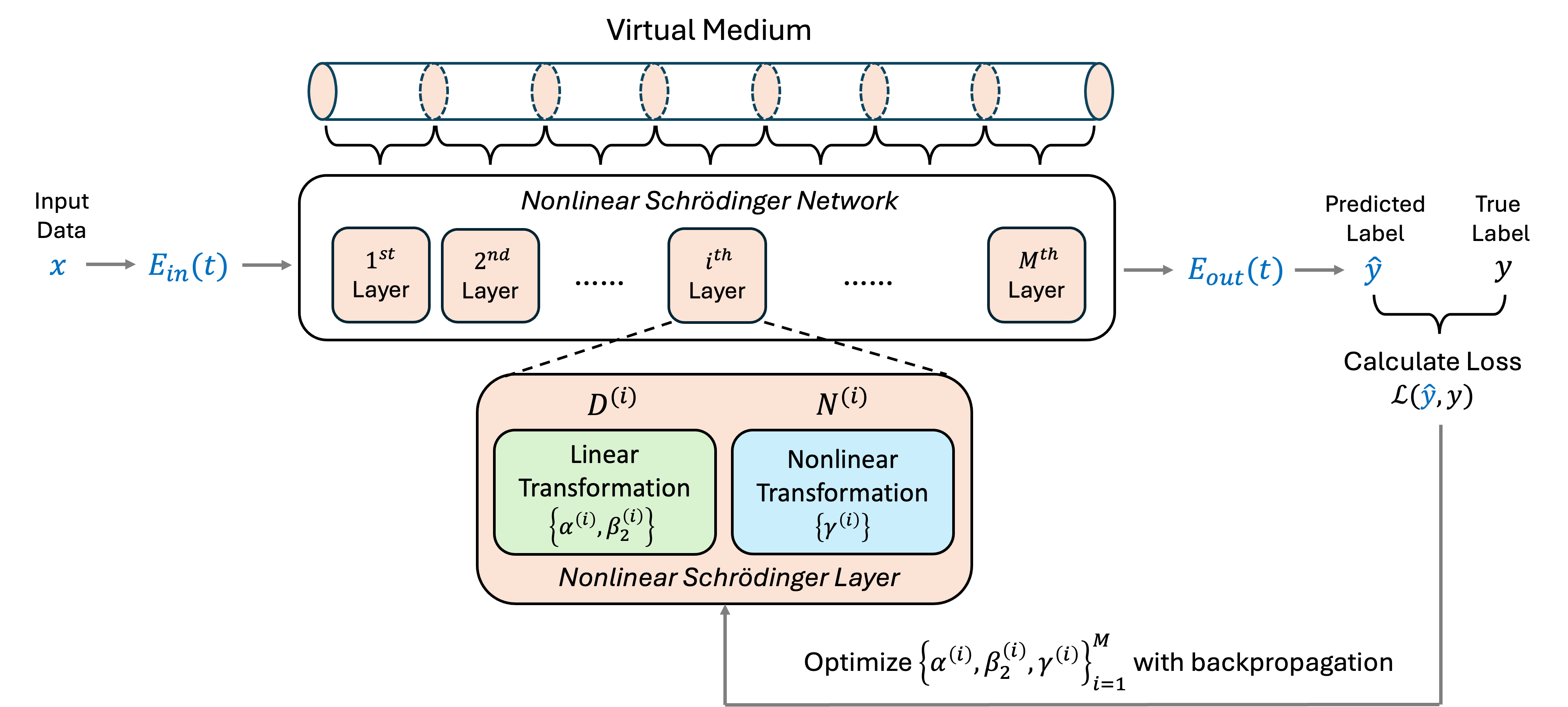}
    \caption{Schematic representation of the Nonlinear Schrödinger Network. The input data $x$ is treated as an input $E_{in}(t)$ propagating through a virtual medium. The network consists of $M$ cascaded Nonlinear Schrödinger Layers, each comprising a linear transformation parameterized by $\alpha$ and $\beta_2$, followed by a nonlinear transformation parameterized by $\gamma$. These layers transform the input according to the Nonlinear Schrödinger Equation (NLSE), resulting in the transformed output  $E_{out}(t)$. The output is then mapped to the predicted label $\hat{y}$. The trainable parameters $\{\alpha, \beta_2, \gamma\}$ across all the layers are optimized through backpropagation and gradient descent to minimize the loss between the predicted and true labels.}
    \label{fig:schrodinger_network}
\end{figure*}

Traditionally, the Nonlinear Schrödinger Equation is numerically solved using the split-step Fourier method, which discretizes the propagation medium into alternating linear and nonlinear operators \cite{agrawal2000nonlinear}. This solution approach lays the groundwork for the Nonlinear Schrödinger Network, wherein we optimize the parameters of these linear and nonlinear transformations to embed data in a trainable physics-based model. Figure~\ref{fig:schrodinger_network} presents a schematic diagram of the Nonlinear Schrödinger Network. The input data $x$ is treated as an input $E_{in}(t)$ that propagates through a virtual medium, wherein the propagation over a short distance is represented by a Nonlinear Schrödinger Layer with two main components: a linear transformation $D^{(i)}$ parameterized by $\alpha^{(i)}$ and $\beta_2^{(i)}$, followed by a nonlinear transformation parameterized by $\gamma^{(i)}$, as described below.

\begin{equation}
  D^{(i)}(E(t)) = \mathcal{F}^{-1} \left\{ \mathcal{F}\{E(t)\} \cdot \exp\left(- \frac{\alpha^{(i)}}{2} \delta z - i\frac{\beta_2^{(i)}}{2} \omega^2 \delta z \right) \right\}
  \label{eq:nlse_D}
\end{equation}

\begin{equation}
  N^{(i)}(E(t)) = E(t) \cdot \exp\left(i\gamma^{(i)} |E(t)|^2 \delta z\right)
  \label{eq:nlse_N}
\end{equation}
\newline

This structure resembles a single block in a common neural network, where a linear layer is typically followed by a nonlinear activation function. By cascading several Nonlinear Schrödinger Layers, the Nonlinear Schrödinger Network is capable of learning nonlinear mappings from the input data to the desired output.

In Equation \ref{eq:nlse_D} and \ref{eq:nlse_N}, $E(t)$ is a 1D vector with $\omega$ being its corresponding frequency vector. $\alpha^{(i)}$, $\beta_2^{(i)}$, and $\gamma^{(i)}$ are trainable scalars in each layer, and $\delta z$ is a constant scalar. $\mathcal{F}$ and $\mathcal{F}^{-1}$ represent 1D Fourier transform and 1D inverse Fourier transform, respectively. As the input data traverses the layers, it undergoes a series of transformations governed by the NLSE, resulting in the transformed output $E_{out}(t)$. Such an output is then mapped to the predicted label $\hat{y}$, which is compared with the true label $y$ to compute the loss $\mathcal{L}$. With all the operations being differentiable, backpropagation and stochastic gradient descent are utilized to update the trainable parameters $\{\alpha^{(i)}, \beta_2^{(i)}, \gamma^{(i)}\}_{i=1}^M$ across all the layers. 

Remarkably, each Nonlinear Schrödinger Layer only introduces three trainable parameters, yet they can parameterize the nonlinear transformation more efficiently than conventional neural networks, as demonstrated later in this work. Moreover, since the trainable parameters are coefficients of a physics equation, the resulting model is interpretable by the physics. This unique property distinguishes the Nonlinear Schrödinger Network from traditional black-box neural networks.

It is worth noting that we modulate the original 1D data onto a super-Gaussian pulse with zero padding for more accurate spectral representations during the computation~\cite{agrawal2000nonlinear}. The computation across all the Nonlinear Schrödinger Layers doesn't change the dimension of data, and the desired output $y$ may have different dimensions compared to $E_{out}(t)$, especially in classification tasks. To align with the output dimension, a densely connected linear layer will be added.

\section{Results}

\subsection{Time Series Classification}

We demonstrate the performance of the proposed Nonlinear Schrödinger Network on three time series classification datasets from different domains and compare it with conventional neural networks.

\begin{itemize}
    \item \textbf{Ford Engine Dataset}~\cite{dau2019ucr}: This dataset consists of time series representing measurements of engine noise. The classification problem is to diagnose whether a specific symptom exists or not in an automotive subsystem, which is a binary classification task.
    \item \textbf{Starlight Curve Dataset}~\cite{rebbapragada2009finding}: Each time series in this dataset records the brightness of a celestial object as a function of time. The classification task is to associate the light curves with three different sources.
    \item \textbf{Spoken Digits Dataset}~\cite{jackson2018jakobovski}: This dataset contains recordings of digits (0-9) at 8kHz from different speakers. The classification task is to classify the recording as one of the ten digits.
\end{itemize}

\begin{figure*}
  \centering
  \includegraphics[width=5in]{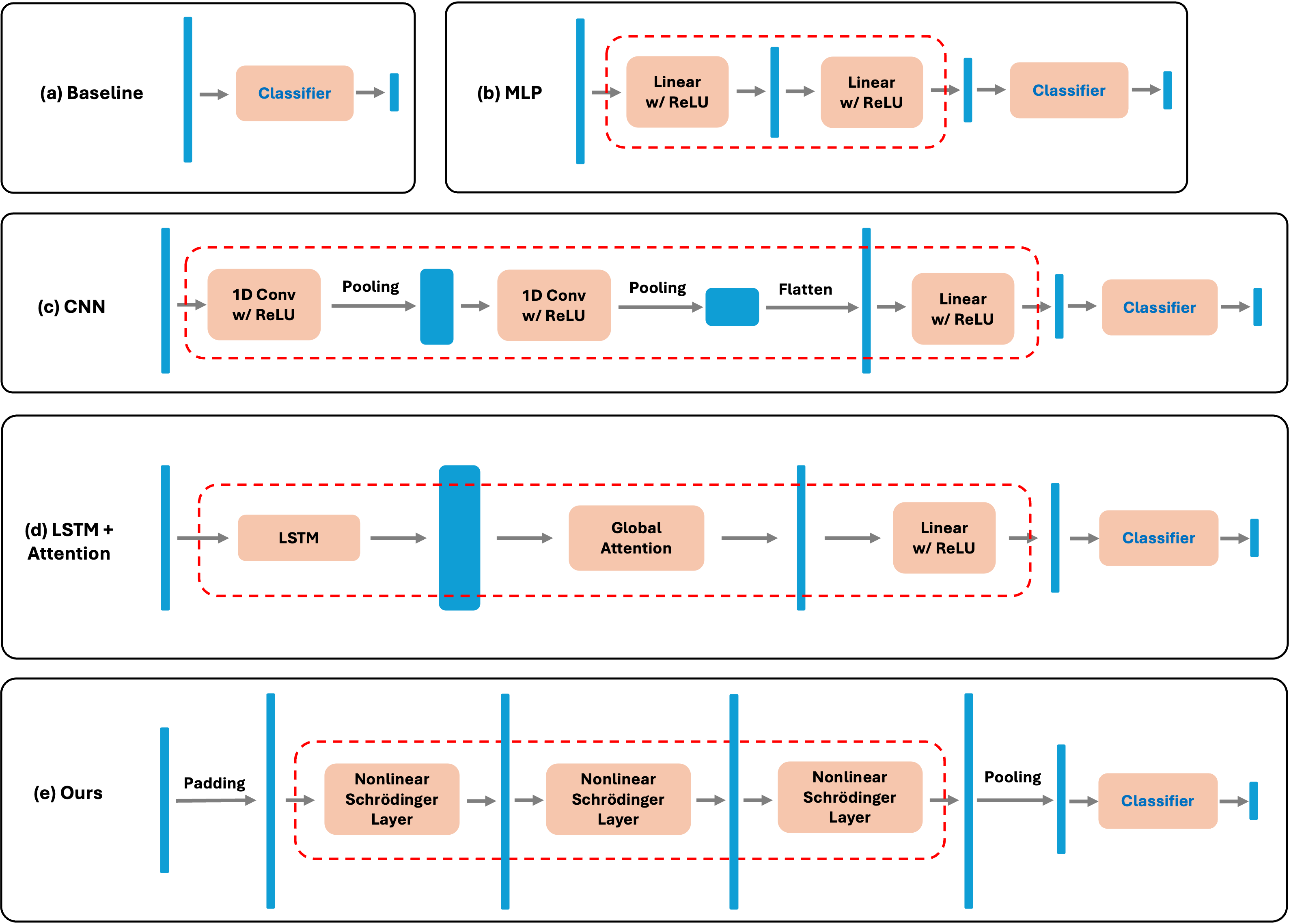}
  \caption{Model architectures of (a) baseline (linear classifier), (b) Multi-Layer Perceptron (MLP), (c) Convolutional Neural Network (CNN), (d) Long Short-Term Memory (LSTM) with Attention:.(d) Nonlinear Schrödinger Network. The evolutions of data dimensions are visualized in blue rectangles. The ``embedding'' layers are highlighted with the red dashed box.}
  \label{fig:models_compare}
\end{figure*}

Table~\ref{tab:nlse-benchmark} presents the accuracy and the number of trainable parameters among four different models across the three datasets. The model architectures are visualized in Figure~\ref{fig:models_compare}. The baseline is a linear classifier consisting of a single densely connected layer. The Multi-Layer Perceptron (MLP) is a feedforward neural network with several densely connected hidden layers and nonlinear activations. The 1D Convolutional Neural Network (CNN) includes a few convolutional layers cascaded with densely connected linear layers. The Long Short-Term Memory (LSTM) model in Figure~\ref{fig:models_compare} is a recurrent neural network that processes input sequences using a 1-layer LSTM. A learned global attention mechanism is used to highlight the most relevant time steps, with the resulting attention scores are normalized via softmax and used to form a single context vector (via a weighted sum of hidden states). After the attention is applied, the output is cascaded through densely connected linear layers. In our proposed Nonlinear Schrödinger Network, we add a max pooling layer after the cascaded Nonlinear Schrödinger Layers to mitigate the dimension increase caused by the zero-padding, and a densely connected linear layer to match the output dimension to the number of classes. All models are trained with the Adam optimizer using the cross-entropy loss. As shown in Figure~\ref{fig:models_compare}, we divide the model architecture into two parts: the ``embedding'' layers (red dashed box) that maps the original data into a more linearly separable space, and a linear classifier that makes the final decision on the class to which the transformed data belongs. Note that there is no embedding layer in the baseline model.

\begin{table}[h]
  \caption{Classification accuracy and number of trainable parameters of different models.}\label{tab:nlse-benchmark}
  \begin{tabular*}{\textwidth}{@{\extracolsep\fill}lcccccc}
  \toprule%
  & \multicolumn{2}{@{}c@{}}{\textbf{Starlight}} & \multicolumn{2}{@{}c@{}}{\textbf{Ford Engine}} & \multicolumn{2}{@{}c@{}}{\textbf{Spoken Digits}}
  \\\cmidrule{2-3}\cmidrule{4-5}\cmidrule{6-7}%
  Metrics & Param \# & Acc (\%) & Param \# & Acc (\%) & Param \# & Acc (\%) \\
  \midrule
  Baseline  &  3,072  & 84.6 (0.6) & 500        & 49.1 (0.5) & 51,200       & 15.6 (1.1) \\
  MLP   & \textcolor{red}{139,264}+192 & 85.8 (0.7) & \textcolor{red}{72,385}+64  & 87.7 (0.7) & \textcolor{red}{663,754}+640  & 23.1 (1.6) \\ 
  CNN   & \textcolor{red}{65,715}+96   & 92.0 (1.9) & \textcolor{red}{32,153}+32  & 89.5 (1.8) & \textcolor{red}{327,866}+320  & 61.9 (8.4) \\ 
  LSTM+Attention & \textcolor{red}{91,969}+195 & 86.7 (0.7) & \textcolor{red}{23,457}+66 & 90.9 (0.5) & \textcolor{red}{364,161}+1,290 & 54.4 (6.5) \\
  Ours  & \textcolor{red}{18}+3,072  & 92.1 (1.5) & \textcolor{red}{18}+500   & 86.2 (3.3) & \textcolor{red}{18}+51,200  & 70.1 (2.5) \\
  \botrule
  \end{tabular*}
  \footnotetext{Baseline: a single layer linear classifier. MLP: Multi-Layer Perceptron. CNN: 1D Convolutional Neural Network. LSTM+Attention: 1-layer Long Short-Term Memory Recurrent Neural Network with Global Attention. Ours: Nonlinear Schrödinger Network. The number of trainable parameters in the ``embedding'' layers are highlighted in red. All reported accuracies are averaged over 10 random seeds, with standard deviation in parentheses.}
\end{table}

The baseline results obtained using only a linear classifier demonstrate poor performance, particularly on linearly non-separable datasets such as Ford Engine and Spoken Digits. This highlights the inherent limitations of linear classifiers in capturing complex patterns and relationships in the data. In contrast, our proposed Nonlinear Schrödinger Network significantly improves the classification accuracy by incorporating several Nonlinear Schrödinger Layers as embedding layers preceding the linear classifier. Each added layer only introduces 3 extra parameters $\{\alpha^{(i)}, \beta_2^{(i)}, \gamma^{(i)}\}$ to the model as discussed earlier, efficiently parameterizing the nonlinear transformation that projects the original data into a more linearly separable space, thereby improving the performance of the same linear classifier used in the baseline model. The results reported here for the Nonlinear Schrödinger Network utilize 6 layers for all three datasets. Therefore, the Nonlinear Schrödinger Network introduces only 18 additional trainable parameters compared to the baseline model as highlighted in the Table. This is a notable advantage of the Nonlinear Schrödinger Network, as it can achieve high performance with a minimal increase in model complexity.

We further compare the Nonlinear Schrödinger Network with conventional neural networks, including Multilayer Perceptrons (MLPs), 1D Convolutional Neural Networks (CNNs), and a Long-Short Term Memory (LSTM) recurrent neural network with global attention. As shown in Table~\ref{tab:nlse-benchmark}, the Nonlinear Schrödinger Network achieves comparable or better accuracy while maintaining significantly fewer parameters across time series classification results. The moderate performance of the LSTM model relative to our Nonlinear Schrödinger Network is consistent with observed patterns in time series classification literature. Recurrent Neural Networks (RNNs), including LSTMs, are fundamentally designed for sequential output prediction rather than classification of entire sequences, often struggle with gradient propagation despite their gating mechanisms, and present significant training and parallelization challenges \cite{ismail2019deep}. These architectural limitations likely explain why the LSTM model, despite its substantially larger parameter count, fails to demonstrate consistent performance advantages over our proposed approach across the evaluated datasets.

Notably, the number of trainable parameters in the embedding layers of the Nonlinear Schrödinger Network is orders of magnitude fewer than the other models, as highlighted in red. These results demonstrate the effectiveness of the proposed Nonlinear Schrödinger Layers in capturing the complex nonlinear dynamics of time series data while maintaining a compact model size. 

\subsection{Learning Curve and Parameter Convergence}

\begin{figure*}
    \centering
    \includegraphics[width=4in]{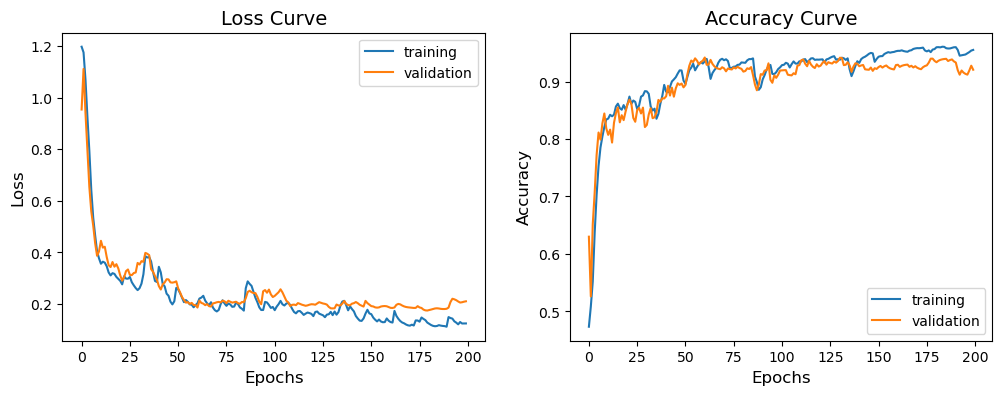}
    \caption{Learning curves of the Nonlinear Schrödinger Network trained on the Starlight dataset. The progression of loss (left) and accuracy (right) for both training and validation datasets over the training epochs are demonstrated here. The curves have been smoothed using a weighted moving average to enhance visual clarity.}
    \label{fig:loss_acc}
\end{figure*}

\begin{figure*}
  \centering
  \includegraphics[width=5in]{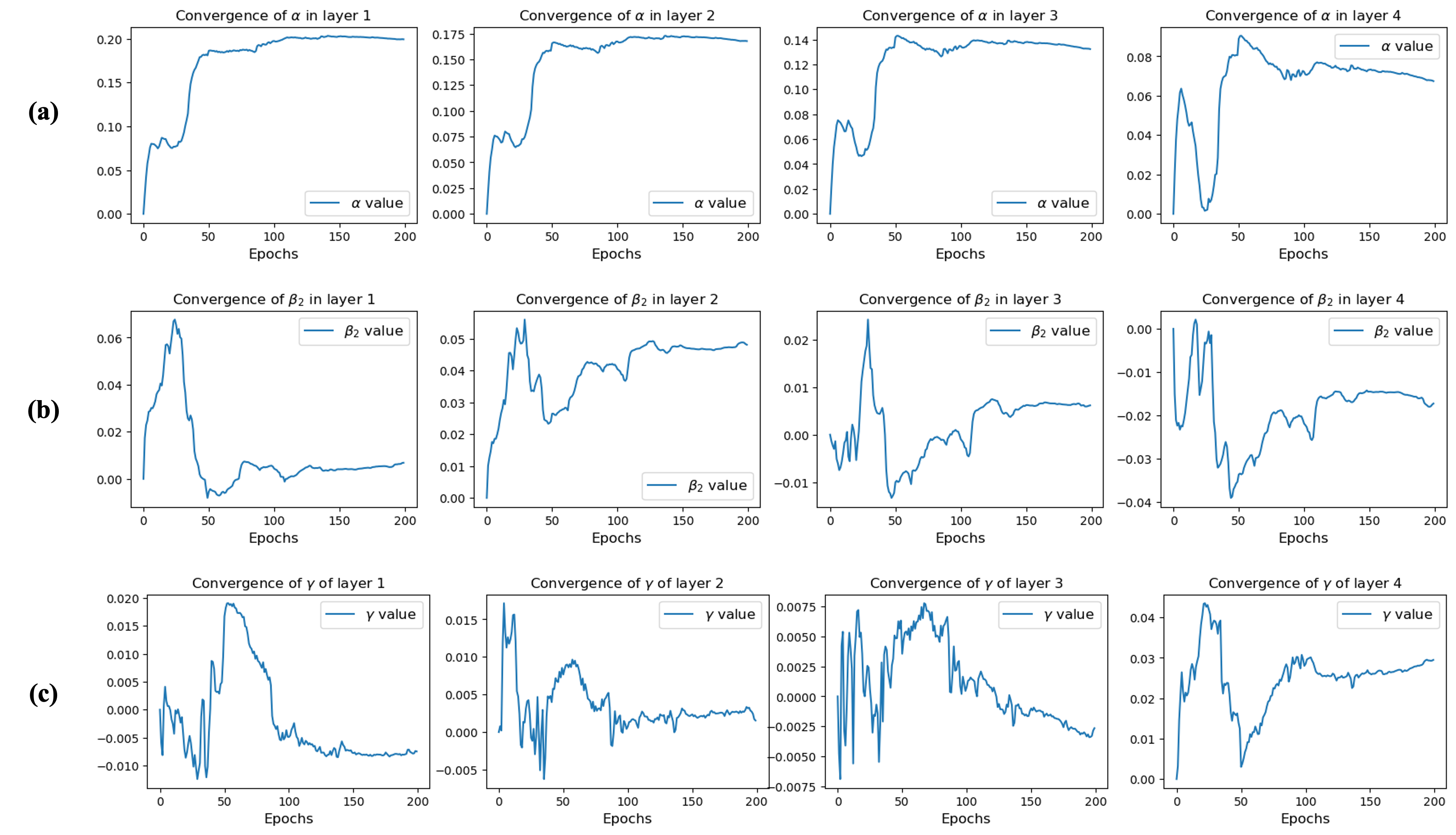}
  \caption{The convergence of $\alpha$ or attenuation, $\beta_2$ or group delay dispersion,  and $\gamma$ or nonlinearity, of the physical system in the first four layers of the Nonlinear Schrödinger Network during training on the Starlight dataset. (a), (b), and (c) depict the convergence of $\alpha, \beta_2,$ and $\gamma$, respectively. All parameters start with zero and gradually converge to values that optimize the network's classification performance.}
  \label{fig:param_converge}
\end{figure*}

Figure~\ref{fig:loss_acc} illustrates the learning curves of the Nonlinear Schrödinger Network trained on the Starlight dataset. As the training progresses, the loss decreases steadily while the accuracy improves for both the training and validation datasets. This indicates successful optimization of the network using backpropagation, demonstrating the network's ability to learn and generalize from the training data.

Figure~\ref{fig:param_converge} presents the convergence of the key parameters $\alpha, \beta_2$, and $\gamma$ in the first four layers of the Nonlinear Schrödinger Network during training on the Starlight dataset. Initially, all parameters are set to zero. As the training progresses, these parameters gradually converge to values that optimize the network's classification performance. Notably, they can be interpreted as the properties of a virtual physical system that transforms the original data to another space for improved classification accuracy.

\subsection{Ablation Study on the Impact of Dispersion and Nonlinearity}

In this section, we conduct an ablation study to investigate the effects of the linear operations ($D$), as shown in Equation~\ref{eq:nlse_D}, and the nonlinear operations ($N$), as shown in Equation~\ref{eq:nlse_N}, within the Nonlinear Schrödinger Network. This study aims to provide insights into the network's underlying mechanisms and how these components interact to achieve improved performance. 

To accomplish this, we first train the network using only the linear operations ($D$) by setting $\gamma^{(i)}=0$ and updating only $\{\alpha^{(i)}$ and $\beta_2^{(i)}\}$ across all layers during the training process. Similarly, we then train the network using only the nonlinear operations ($N$) by setting $\{\alpha^{(i)}=0, \beta_2^{(i)}=0\}$ and updating only $\gamma^{(i)}$ across all layers during training. The results of these ablated models are compared with the baseline (a linear classifier only) and the full Nonlinear Schrödinger Network (with both linear and nonlinear effects enabled) in Table~\ref{tab:ablation}.

\begin{table}[h]
  \caption{Ablation study of linear and nonlinear operations in the Nonlinear Schrödinger Layer.}\label{tab:ablation}%
  \begin{tabular}{@{}lcc@{}}
  \toprule
  Model &  Ford Engine Acc (\%) &  Spoken Digits Acc (\%) \\
  \midrule
  Baseline (Linear Classifier Only) & 49.1 (0.5) & 15.6 (1.1)  \\
  Nonlinearity ($N$) Only & 51.7 (1.2) & 19.9 (1.9)  \\
  Dispersion ($D$) Only & 79.5 (6.2) & 60.3 (5.5)  \\ 
  Full ($N$ and $D$)    & 86.2 (3.3) & 70.1 (2.5)  \\ 
  \botrule
  \end{tabular}
  \footnotetext{Baseline: model with a linear classifier. $N$ Only: Nonlinear Schrödinger Network with only nonlinear operations in each layer. $D$ Only: Nonlinear Schrödinger Network with only linear operations in each layer. Full: Both linear and nonlinear operations exist.}
\end{table}

The results demonstrate that the model with only linear operations ($D$) achieves significant improvement over the baseline. In contrast, the model with only nonlinear operations ($N$) shows barely any improvement over the baseline. The full model, incorporating both linear and nonlinear components, yields the best results overall. 

These findings can be attributed to the data processing method employed in the Nonlinear Schrödinger Network. The input data is amplitude modulated onto the real input $E_{in}(t)$. After propagation through the network, a complex output $E_{out}(t)$ is obtained, and its amplitude $|E_{out}(t)|$ is subsequently fed to the linear classifier. When only the nonlinear operations ($N$) are present, the self-phase modulation does not alter the amplitude, resulting in $|E_{out}(t)|=E_{in}(t)$. Consequently, the nonlinear operations alone do not significantly contribute to the network's performance. On the other hand, when only linear operations ($D$) are present, all transformations are linear, except for the final step of taking the amplitude. This allows the linear operations to capture some underlying patterns in the data, leading to improved performance over the baseline.

The true potential of the Nonlinear Schrödinger Network is realized when both linear and nonlinear components are combined. The nonlinear effect of $N$ in layer $i$ comes into play when followed by a linear operation $D$ in layer $i+1$ as the data remains in the complex domain. This coupled interaction between the nonlinear and linear components enables the network to capture complicated patterns and relationships in the data more effectively, resulting in the best overall performance.

\section{Extension to Gross-Pitaevskii Equation}

The mathematical framework established for the Nonlinear Schrödinger Network can be generalized to other partial differential equations, demonstrating the broader applicability of the physical embedding paradigm. A natural extension is to the Gross-Pitaevskii equation (GPE), which serves as a mean-field approximation for the quantum dynamics of dilute boson gases, particularly in describing Bose-Einstein condensates at ultra-low temperatures \cite{bao2003numerical}. The GPE is structurally very similar to the NLSE with the inclusion of an external potential term, thereby providing enhanced capabilities for modeling spatially confined quantum systems. 

To explicitly establish the mapping between our NLSE implementation and the GPE framework, we formulate the Hamiltonian of the GPE as:

\begin{equation}
i\hbar\frac{\partial\psi(\mathbf{r},t)}{\partial t} = \left[-\frac{\hbar^2}{2m}\nabla^2 + V(\mathbf{r}) + g|\psi(\mathbf{r},t)|^2\right]\psi(\mathbf{r},t)
\end{equation}
\newline

When mapping between these equations, we establish the following parameter correspondences: the dispersion term $\beta_2$ in our NLSE framework corresponds to $-\hbar^2/m$ in the GPE formulation, while the nonlinearity coefficient $\gamma$ maps to $-g$, where $g$ characterizes interparticle interactions in the condensate. Conversely, the GPE introduces a potential term $V(\mathbf{r})$ that captures spatial confinement effects absent in the original NLSE. This term becomes an additional trainable parameter in our network architecture, complementing the existing parameter set and enhancing the model's representational capacity. The implementation of the GPE within our network architecture requires a structural modification to the layer composition. Each Gross-Pitaevskii Layer is characterized by an augmented parameter set: $\{\beta_2^{(i)}, \gamma^{(i)}, V^{(i)}\}$, where $V^{(i)}$ parametrizes the external potential function. The potential transformation can be parameterized as:

\begin{equation}
  V^{(i)}(E(t)) = E(t) \cdot \exp\left(-iV(t)\delta z\right)
  \label{eq:nlse_N}
\end{equation}
\newline

The potential function in this case is parameterized by a learned polynomial function, with polynomial order considered a hyperparameter, in our case kept at 20. Table \ref{tab:gpe_ablation} presents an ablation analysis comparing the performance metrics of various configurations of the GPE-based network against the baseline linear classifier. The results suggest that the inclusion of the potential term $V$  enhances the model's capacity over the Nonlinear Schrödinger Network when the linear operation $D$ is included but the nonlinear $N$ is excluded. The dataset-specific performance improvements observed with the GPE implementation underscore the ability of the potential term to replace the nonlinearity term in the Nonlinear Schrödinger Network formulation.

\begin{table}[h]
\centering
\caption{Ablation study on the impact of potential term from Gross-Pitaevskii equation.}
\label{tab:gpe_ablation}
\begin{tabular}{lcc}
\hline
Model & Ford Engine Acc (\%) \\
\hline
Baseline (Linear Classifier Only) & 49.1 (0.5) \\
Nonlinearity ($N$) and Potential Energy ($V$) Only & 52.1 (0.1) \\
Dispersion ($D$) and Potential Energy ($V$) Only & 85.0 (7.6) \\
Full ($N$ and $D$ and $V$) & 86.0 (2.4) \\
\hline
\end{tabular}
\footnotetext{Baseline: model with a linear classifier. Nonlinearity $N$ and Potential $V$ Only: GPE Network with only nonlinear and potential operations in each layer. Dispersion $D$ and Potential $V$ Only: GPE Network with only dispersive and potential operations in each layer. Full: Dispersive, nonlinear, and potential terms are all utilized.}
\end{table}

\section{Discussion}

The experiments on various time series classification datasets have demonstrated that the Nonlinear Schrödinger Network achieves comparable or superior accuracy with orders of magnitude fewer parameters than conventional models such as Multi-Layer Perceptrons (MLPs) and Convolutional Neural Networks (CNNs). This highlights the efficiency of our approach in capturing complex dynamics with minimal model complexity. Furthermore, the transformations learned by the Nonlinear Schrödinger Network can be explicitly formulated and interpreted through the lens of a partial differential equation, with each parameter having a clear physical meaning. This contrasts sharply with the black-box nature of typical neural networks, offering insights into the data transformation process and making the model more transparent and understandable. 

The extension to the GPE demonstrates that our approach is not limited to the NLSE but constitutes a general framework that can be adapted to various PDEs that model different physical phenomena. The success of this extension suggests promising avenues for incorporating other master equations of physics into trainable computational models.

\subsection{Implementation in Analog Optical Computing}

Given the physical underpinnings of the approach, there is potential for analog implementation, where the ultrafast timescales of pulse propagation can drastically reduce latency compared to traditional digital systems \cite{solli2015analog}. While constructing a system that implements a layer-by-layer NSN poses considerable engineering challenges, a new concept in AI optical hardware acceleration offers a promising alternative \cite{zhou2022nonlinear}. This method modulates data onto a supercontinuum laser spectrum, which then undergoes a nonlinear optical transformation analogous to a kernel operation, enhancing data classification accuracy. It is shown that the nonlinear optical kernel can improve linear classification results similar to a traditional numerical radial-basis-function (RBF) kernel but with orders of magnitude lower latency. The inference latency of the RBF kernel with a linear classifier is on the order of $10^{-2}$ to $10^{-3}$ second, while the nonlinear Schrödinger kernel achieves a substantially reduced latency on the order of $10^{-5}$ seconds and better classification accuracy. 

In the original work, the performance is data-dependent due to the limited degrees of freedom and the unsupervised nature of the optical kernel. A follow-up work has demonstrated that by modulating the spectral phase of the input data, the nonlinear optical interactions within the kernel can be effectively trained in a data-driven manner \cite{zhou2023low}. This approach exploits a similar physical embedding process, with the same dispersive and nonlinear effects, reinterpreted as adjustable spectral-phase manipulations. Through this process, data-driven training and inference becomes feasible in a physically constrained, ultrafast domain. Such approaches provide potential groudwork for analog implementation of the Nonlinear Schrödinger Network.

\subsection{Generalizability}

While the Nonlinear Schrödinger Network demonstrates impressive performance, it is essential to acknowledge that its generalizability might be less than that of traditional DNNs. The model's design is heavily influenced by the specific physical principles encoded in the NLSE. As a result, it may not be as flexible in handling data that do not conform to the underlying physical assumptions. However, this trade-off between interpretability and generalizability is not necessarily a disadvantage. In many practical applications, a more physically consistent and interpretable model may be preferred, even if it sacrifices some degree of generalizability.

\subsection{Other Limitations}

A natural question is how broadly this idea applies to datasets or tasks that do not exhibit wave-like or dispersive phenomena. Since not all data domains align naturally with the physical assumptions underlying NLSE (or even PDEs in general), it raises questions about how effective ``physical embedding'' would be for, say, image-based tasks, high-dimensional tabular data, or domains with more complex spatiotemporal patterns. More varied benchmarking (particularly comparing to state-of-the-art methods on widely used benchmark corpora) would also help establish how well the approach generalizes beyond relatively moderate-sized data.

\section{Conclusion}

In this paper, we introduced the concept of physical embedding and demonstrated it through the Nonlinear Schrödinger Network (NSN), which repurposes the Nonlinear Schrödinger Equation (NLSE) as a trainable model for learning complex data mappings. Instead of relying on large sets of opaque parameters, the NSN embeds representations in a compact parameter space governed by physically interpretable coefficients. This approach, which directly encodes the data into a set of fundamental wave-equation properties, offers improved interpretability and drastically reduced memory requirements.

Our experiments on time series classification show that the NSN performs on par with, or surpasses, standard deep-learning architectures—yet requires orders of magnitude fewer parameters. Ablation studies highlight that its synergy of dispersion and nonlinearity is key to modeling intricate temporal patterns. Moreover, the concept of physical embedding readily adapts to other partial differential equations such as the Gross-Pitaevskii Equation, underscoring its broader relevance wherever domain-specific physics can be leveraged.

In addition to these gains in parameter efficiency, analog implementation stands out as an exciting future direction. By physically realizing dispersion and nonlinear effects in hardware—particularly in optical platforms—the NSN could perform ultrafast inference with significantly lower energy consumption. While there are trade-offs in terms of domain specificity and flexibility, the prospect of physical embedding offers a powerful alternative to black-box networks, promising deeper interpretability, resource-conscious scaling, and new pathways for physics-based machine learning.
\backmatter

\bibliography{main}

\end{document}